\documentclass[sigconf]{acmart}
\pdfoutput=1

\AtBeginDocument{%
  \providecommand\BibTeX{{%
    \normalfont B\kern-0.5em{\scshape i\kern-0.25em b}\kern-0.8em\TeX}}}

\usepackage{todonotes}
\usepackage{hyperref}

\graphicspath{{./pdf/}{./jpeg/}{./images/}}
 \DeclareGraphicsExtensions{.pdf,.jpeg,.jpg,.png}

\begin{document}


\title[Improving Borderline Adulthood Facial Age Estimation]{Improving Borderline Adulthood Facial Age~Estimation~through~Ensemble~Learning}





\settopmatter{authorsperrow=4}



\author{Felix Anda}
\authornote{UCD Forensics and Security Research Group - \url{https://www.forensicsandsecurity.com}}
\email{felix.anda@ucdconnect.ie}
\affiliation{%
  \institution{University College Dublin}
  \city{Dublin}
  \country{Ireland}
}

\author{David Lillis}
\authornotemark[1]
\email{david.lillis@ucd.ie}
\affiliation{%
  \institution{University College Dublin}
  \city{Dublin}
  \country{Ireland}
}

\author{Aikaterini Kanta}
\authornotemark[1]
\email{aikaterini.kanta@ucdconnect.ie}
\affiliation{%
  \institution{University College Dublin}
  \city{Dublin}
  \country{Ireland}
}

\author{Brett A. Becker}
\authornotemark[2]
\email{brett.becker@ucd.ie}
\affiliation{%
  \institution{University College Dublin}
  \city{Dublin}
  \country{Ireland}
}

\author{Elias Bou-Harb}
\email{ebouharb@fau.edu}
\authornote{Cyber Threat Intelligence Laboratory}
\affiliation{%
  \institution{Florida Atlantic University}
  \city{Boca Raton, FL}
  \country{USA}
}

\author{Nhien-An Le-Khac}
\authornotemark[1]
\email{an.lekhac@ucd.ie}
\affiliation{%
  \institution{University College Dublin}
  \city{Dublin}
  \country{Ireland}
}

\author{Mark Scanlon}
\authornotemark[2]
\email{mark.scanlon@ucd.ie}
\affiliation{%
  \institution{University College Dublin}
  \city{Dublin}
  \country{Ireland}
}


\renewcommand{\shortauthors}{Anda et al.}

\begin{abstract}

Achieving high performance for facial age estimation with subjects in the borderline between adulthood and non-adulthood has always been a challenge. Several studies have used different approaches from the age of a baby to an elder adult and different datasets have been employed to measure the mean absolute error (MAE) ranging between 1.47 to 8 years. The weakness of the algorithms specifically in the borderline has been a motivation for this paper. In our approach, we have developed an ensemble technique that improves the accuracy of underage estimation in conjunction with our deep learning model (DS13K) that has been fine-tuned on the Deep Expectation (DEX) model. We have achieved an accuracy of 68\% for the age group 16 to 17 years old, which is 4 times better than the DEX accuracy for such age range. We also present an evaluation of existing cloud-based and offline facial age prediction services, such as Amazon Rekognition, Microsoft Azure Cognitive Services, How-Old.net and DEX.

\end{abstract}


\copyrightyear{2019} 
\acmYear{2019} 
\setcopyright{acmlicensed}
\acmConference[ARES '19]{Proceedings of the 14th International Conference on Availability, Reliability and Security (ARES 2019)}{August 26--29, 2019}{Canterbury, United Kingdom}
\acmBooktitle{Proceedings of the 14th International Conference on Availability, Reliability and Security (ARES '19), August 26--29, 2019, Canterbury, United Kingdom}
\acmPrice{15.00}
\acmDOI{10.1145/3339252.3341491}
\acmISBN{978-1-4503-7164-3/19/08}



\maketitle

\section{Introduction}
\label{intro}
Automated facial age estimation is the application of non-manual processes to measure the age of a person by analysing specific facial features with the use of artificial intelligence. 
Facial age related products are becoming increasingly popular in our daily life. Teenagers and adults are using ageing filters for entertainment purposes that are available with Snapchat which have become viral over the past years. These new face-ageing techniques could boost search for wanted criminals or missing people. The usage of facial recognition has incremented exponentially. Furthermore, biometric systems are expanding their robustness with the addition of facial-based authentication factors that prevent impersonation attacks, e.g., Apple's Face ID and Android's face recognition technologies.

Facial recognition is a widely-used technology that maps facial features from images to detect faces and recognise the associated identity. Applications have been commonly found in airports, mobile devices and certain web pages~\cite{5406526}. Entertainment venues, alcohol, tobacco and certain social media services require an age verification process. Facial recognition is shaping the future of several security innovations: facial security checks could be used to prevent credit card cloning, smartphone unauthorised access, fraudulent exam takers, fake social media accounts, etc. Facial age detection could also be used to prevent unauthorised consumption or purchase of certain goods or services. Undocumented criminals are open to deceive authorities about their age to avoid the judicial system; however, an automated age detector could impede their attempt to bypass the system.

Accurate facial age estimation has long been a difficult task for both human experts and specialised machine learning algorithms. Moreover, the influence of factors, such as environment, health habits, lifestyle, makeup, emotions, and uncontrolled lightning hinder the age estimation process~\cite{han2013age}. We have studied the possibility of including artificial intelligence as a means to detect and analyse evidence that may be presented in court. Specifically, we have focused on the improvement of facial age estimation algorithms for the identification of victims/suspects and its applications to child sexual exploitation material (CSEM) and child sexual abuse material (CSAM) investigations\footnote{These are the terms recommended by the Luxembourg Guidelines\\ (\url{http://luxembourgguidelines.org/})}. Challenges arise due to the factors previously mentioned, thus hampering age classification accuracy, especially for borderline cases between underage and adult subjects. Due to the nature of courtroom practice, and the necessity of expert testimony, it is neither intended nor anticipated that these AI techniques will fully replace trained investigators. Rather, this type of investigative aid has the potential to greatly expedite digital forensic analysts in their work, and potentially lower the psychological load of dealing with CSEM material on an ongoing basis.

The usage of Deep Learning in several fields has become the latest trend: at the end of 2018, Google introduced an AI tool (freely available for non-governmental organisations and industry partners) to assist organisations in detecting and reporting child sexual abuse material online~\cite{Google2018}. With the emergence of AI and its state-of-the-art branches including computer vision, machine learning and deep learning, age determination has improved significantly. Neural networks learn by processing thousands of images so that they can predict the age of future unseen images at an accuracy that surpasses human facial age perception capacities.

Given the quantity of digital content being created daily, the previous approach of manual evidence analysis is unfeasible~\cite{le2018deeplearningmalware}. However, machines require training to deliver accurate estimations. The training process demands a large volume of labelled data, extensive time, and computer resources to understand traits present in digital portraits. In a previous study, several age estimation services were evaluated throughout an age range of 0 to 77 years old~\cite{anda2018evaluating}. With this study it was found that the real culprit of inaccurate age predictions for minors is linked to the lack of appropriate datasets with adequate age labels. Nonetheless, data collection of underage images is surrounded by ethical and moral concerns. Personal identifiable information, such as name, gender, age, and additional information must be handled with care and the exposure of sensitive information by uploading underage images in an unencrypted Internet can be detrimental. Conversely, data collection with the appropriate safeguards could assist missing children and detect previously unknown child abuse material. 

Child exploitation investigations are one of the more common investigation types in digital forensic laboratories throughout the world~\cite{ANDA2019S142}. These investigations have become an arduous task due to the increasing usage of anonymization tools, private P2P networks and cloud-based KVM systems~\cite{farina2015overviewcloudforensics}. Worldwide, law enforcement and child protection communities have been fighting to diminish CSEM and human trafficking. Automated age detection techniques can be used to reduce work exposure to incriminating archives of indecent images; therefore, reducing the psychological ramifications. Such techniques have also been exercised for image classification and categorisation according to age, gender, objects contained therein, and the location in which each image was taken, all of which are useful to CSEM investigators.
 
\subsection{Contribution of this Work}
\label{contribution}
The contribution of this work can be summarised as:

\begin{itemize}

\item Comprehensive performance evaluation of offline and cloud-based facial recognition models.

\item The development and evaluation of a novel deep learning based underage subject classification model, \texttt{DS13K} with N=12792 images, 80\% for training and 20\% for testing.

\item Significant improvement over individual cloud-based age estimators through the use of ensemble-based approaches for subjects under the age of 18 - comparable with expert human estimators.

\end{itemize}

\section{Literature Review/State of the Art}
\label{ageing}

\subsection{Automated Age Estimation}
\label{age_recognition}

The human face can reveal important information, such as gender, approximate age, skin tone, eye colour, hair colour, presence/absence of makeup, presence/absence of beard, presence/absence of moustache, etc. All these elements are know as soft biometric traits. \citet{Dantcheva2011} defines soft biometric traits as ``physical, behavioural or adhered human characteristics, classifiable in predefined human compliant categories''.

Accurately determining the age of a victim can prove crucial in a CSEM possession and/or distribution case, especially for borderline age ranges between underage teenagers and young adults. The prediction of age as a soft biometric trait has been proven to be difficult due to the absence of strong cues that determine the oldness of a subject. \citet{kloess2017challenges} suggest that discrepancies between the face and body, natural variation between different ethnicities and the environment that the person is exposed to are factors that affect the age prediction process. The aforementioned research takes into account multiple factors that can lead to the classification of an image either if it is an indecent image and the respective age group.

The mean absolute error (MAE) and the mean absolute error per age (MAE/A) are the performance metrics used throughout this paper. The former is the average difference between the predicted age and the ground truth; the latter is the MAE grouped by the age.

In the past two decades, error rates have decreased remarkably. A MAE of 1.47 was achieved by \citet{ratnayake2014juvenile} in 2014 by accomplishing an AdaBoost\footnote{AdaBoost is a machine learning boosting algorithm that iteratively builds an ensemble of models~\cite{seiffert2008rusboost}.} fusion of several state-of-the-art classifiers (including Fisher's LDA, Neural Networks, and Support Vector Machine). Nevertheless, this study was executed over a limited private dataset of 50 female images with an age range from 10 to 19, which is indicative of the scarcity of suitable images of this type. In 2011, \citet{luu2011contourlet} were able to obtain a MAE of 4.1 (which has been typical of techniques utilising the FG-NET database). The \textit{contourlet} appearance model used was more accurate and faster at localising facial landmarks than active appearance Models. \citet{ferguson2017juvenile} acknowledged poor accuracy results for age estimation on juvenile faces by human observation. Influence of age, sex and occupation is nullified in the outcome. Moreover, female age estimation was more accurate in younger age groups and male age prediction were more precise after 11 years of age.

\subsection{Transfer Learning}
\label{transfer_learning}

Knowledge transfer, inductive transfer or transfer learning makes use of existent available data to aid the learning on the new target data, which is composed of training and testing~\cite{Dai:2009:EUF:1553374.1553399}.

The use of transfer learning has been increasing throughout the years and has been brought to the attention of researchers where several of them have published pretrained models to assist other researchers and prevent them from executing the tedious task of training data to solve a specific problem. 

Inductive transfer can be beneficial when there is lack of labelled data, copyright issues or when data could be easily outdated. In our study, we are attempting to obtain a sufficient quantity of labelled facial age images; however issues arise due to copyright restrictions, GDPR, and ethical concerns. Therefore, a transfer learning solution is required. In further studies,~\citet{dong2016automatic} exploited the transfer learning strategy to train deep convolutional neural networks from pretrained models due to the scarcity of age labelled face images. Transfer learning is usually expressed through the use of pretrained models, which are simply models created to solve a specific problem and are suitable for re-usability. Less training data is required when successfully transferring a pretrained model to another task.

\subsection{Face Ageing Datasets}
\label{database}

High-quality large-sample-sized facial image datasets annotated with both age and gender are needed to train models that are capable of predicting accurate age. Several age annotated datasets have been released but with certain limitations, such as lack of images in certain age groups, presence of noise in photos that reduce the quality of the dataset and inaccurate age labelling.

IMDB-WIKI is the largest public facial age computer annotated age and gender dataset~\cite{Rothe-IJCV-2016} and has been subject of hundreds of facial recognition studies. The images were scraped from thousands of celebrities in IMDB\footnote{\url{https://www.imdb.com/}} and correlated with Wikipedia\footnote{\url{https://www.wikipedia.org/}}. The collection is quite considerable as the figures reach over half a million; nevertheless, the calculation of age is acknowledged by the authors to not be entirely accurate. We have corroborated that there are inaccurate age labelling and presence of noise. Furthermore, we have taken extra care in using these images due to copyright restrictions.

The FG-NET~\cite{fgnet} dataset contains 82 subjects with photographs of each at varying ages ranging from newborn to 69 years old. Although over 50\% of images in the FG-NET dataset are child images, the demand for underage training and test data has led to the creation of alternative databases. \citet{grd2016creating} produced a private database in 2016 called ageCFBP with a wider age range. In the same year, Boys2Men was released as another private database focused on male child images~\cite{castrillon2016boys2men}. 

MEDS~\citep{founds2011nist} is a mugshot dataset of male and female deceased subjects with the oldness feature annotated but does not contain images of underage individuals. The FERET dataset contains around 14,000 images and is pertinent to face detection~\cite{phillips1998feret}. The age labelling is based solely on human observation.

The OUI-Adience set is a public collection of labelled images obtained by online facial images of Flickr ``in the wild''. Although ~\citet{eidinger2014age} has stated that they use Creative Commons license for their images, we have detected from a sample of 10,842 images, that 89.55\% are associated to images with copyright; therefore, we have avoided the use of such dataset. Another dataset that uses Flickr as a source is the Yahoo Flickr Creative Commons 100M (YFCC100M) that was released in 2014~\cite{thomee2016yfcc100m}. This is the biggest dataset of images and videos publicly available for researchers. Due to the size of the collection and the dataset being distributed solely as the metadata, the database is constantly evolving. (i.e., the photographs need to be downloaded individually from Flickr).

For our studies, a hybrid dataset was created from a variety of those available (IMDB, WIKI, FG-NET, MEDS) using the dataset generator software published by \citet{anda2018evaluating}.

\section{Existing Tools and Models}
\label{existing}

In this section, the current tools for age estimation that are classified in two categories: Offline and Online. For the former, the tools are associated with pretrained models, where the architecture is known and the training dataset may or may not be shared. For the latter, the tools are hosted as cloud services, and the architecture of the neural network and the training dataset are generally unknown.

The main advantage of using an offline pretrained model is that they are usually shared by researchers either in frameworks, such as Caffe\footnote{\url{https://caffe.berkeleyvision.org/}}, Caffe2\footnote{\url{https://caffe2.ai/}}, Keras\footnote{\url{https://keras.io/}} or Pytorch\footnote{\url{https://pytorch.org/}} and thus have no cost. Nonetheless, online tools are associated with machine learning as a service and require a payment per transaction but are much easier to invoke; no installation is required and less local computational power is used.

The age and gender classification using Convolutional Neural Networks (CNNs) is an offline model that was built on the Adience dataset and released in 2005~\cite{levi2015age}. This pretrained model consisted of a CNN architecture that was adapted to work even though the amount of learning data was scarce. Similarly, the ranking CNN for age estimation model was released in 2017 and is also an offline model that is available in the Model Zoo\footnote{\url{https://modelzoo.co/model/using-ranking-cnn-for-age-estimation}}. This model contains a series of basic CNNs that were fine-tuned from the base network trained on the Adience dataset. The result is a binary output and is ultimately added to the final prediction~\cite{Chen_2017_CVPR}.


According to Economy Watch in 2010 \cite{watch2010us}, Amazon acquired ``Rekognition'' from an Artificial Intelligence start-up company from California called Orbeus. The company had developed a facial recognition software that detected traits on images with the use of a library based on Artificial Neural Networks which are computing systems that learn to accomplish tasks by observing examples rather than executing a specific algorithm and are structured by an initial input layer of neurons, one or more hidden layers, and a final layer of output neurons \cite{wang2003artificial}.

The Kairos service has been used for age prediction and face detection; however according to~\citet{anda2018evaluating}, the age estimation performance was lagging behind the rest of the classifiers included in that study. On the contrary, Microsoft Azure Machine Learning is a fully managed cloud service that is powered by a considerable number of machine learning algorithms aimed for scientists, data analysts and developers~\cite{mund2015microsoft}. Per \citet{Weber2016}, it is suggested that Microsoft Azure Cognitive Service uses Multi-layered deep learning technology and is within the top performers for age estimation. Finally, DEX has been subject to hundreds of studies in fields, such as computer vision, deep learning face recognition and age estimation. The huge dataset of over half a million subjects has been used by several researchers and the model has been trained in multiple frameworks, such as Caffe and Keras\footnote{\url{https://github.com/yu4u/age-gender-estimation}}.

Google has not yet released a fully-fledged age estimation service based on image analysis to the public. The Google Vision Cloud API includes facial recognition and facial landmark features, but only allow the recognition of subjects to be categorised as a minor or non-minor and safe search capabilities, such as the recognition of adult content. It could be suggested that the introduction of the Google tool to assist organisations in detecting and reporting child sexual abuse material online previously mentioned in Section~\ref{intro}, is the combination of both the minor/non-minor detector and the adult content detector.

Finally, How-old.net is an application linked to the Microsoft cognitive services and part of Microsoft's \textit{Project Oxford}. In recent years, the tool went viral on social media and was used mainly for entertainment. Today it can be used to predict underage images with a fairly high accuracy as shown in our study.

\section{Dataset Curation for Performance Evaluation}
\label{curation}

In order to perform unbiased experimentation with the four services identified, it was necessary to construct a balanced dataset. Thus, we ensured that there were an equal number of images collected for each age. The dataset generator proposed in~\cite{anda2018evaluating} was used and additional modules for the datasets that are to be discussed in this section were implemented\footnote{\url{https://bitbucket.org/4nd4/image_database}}.

Because the focus of this paper is on the boundary between minority and adulthood, older ages were not considered. Thus, the dataset was limited to an age range of 0 to 25 inclusive. For this dataset, 492 images per age were collected. For younger ages, this quantity of images was not available in existing public dataset, requiring the incorporation of additional manually discovered images. This was achieved by collecting images from Flickr\footnote{Appropriate ethical approval was awarded for this data gathering process from our research institution (University College Dublin)}. Only photos that were available under an appropriate Creative Commons or Public Domain license, and for which accurate age and gender information were available, were considered. The latter information was taken from metadata, such as photo titles, descriptions, or tags. Other images were included from the UTKFace Dataset~\cite{zhifei2017cvpr}. IMDB and WIKI photos were avoided but still used in a low proportion. This dataset was used for the experiment described in Section~\ref{sec:exp1}. Each image is a single frontal face that was cropped and aligned with DLIB\footnote{C++ toolkit containing machine learning algorithms \url{http://dlib.net/}.} with a dimension of 200~x~200 pixels. Each image was processed by a face detector either by the DLIB libraries by using Histogram Oriented Gradients or Convolutional Neural Networks, or the face detection provided by each service discussed in Section~\ref{existing}. Initially we had a collection of 15,000 images but due to non-face recognition, the figure decreased and in order to maintain a balanced dataset, the images had to be reduced to 492 per class hence, we limited the dataset to a total size of 12,792.

\section{Experiments and Results}
\label{results}

Three experiments were conducted and the MAE was calculated with the formula depicted in Equation~\ref{eq:1}, the results of which are presented in the subsections that follows. The first experiment, discussed in Section~\ref{sec:exp1} focused on the wider age range from 0 to 25 years old, to evaluate and compare the four individual services: How-Old.net, AWS, DEX, and Azure. In addition to the services, our deep learning model, \texttt{DS13K} was created. The second experiment involves the evaluation of DS13K. The model performance reached an accuracy of 55.38\% placing it in the top 3 performers after Bagging Regressor and the Gradient Boosting Regressor. The model is described in Section~\ref{sec:exp2}. The final experiment introduces ensemble machine learning techniques to establish whether these will be useful tools to improve upon the performance of the four systems. This is presented in Section~\ref{sec:exp3}.

\begin{equation} \label{eq:1}
MAE = \sum_{i=1}^{n} \frac { \left|predicted_i - real_i\right|} {n}
\end{equation}

\subsection{Underage Range Estimation}
\label{sec:exp1}

\begin{figure*}[h!]
  \centering
  \includegraphics[width=0.82\textwidth,trim={0 0 0 100 cm},clip]{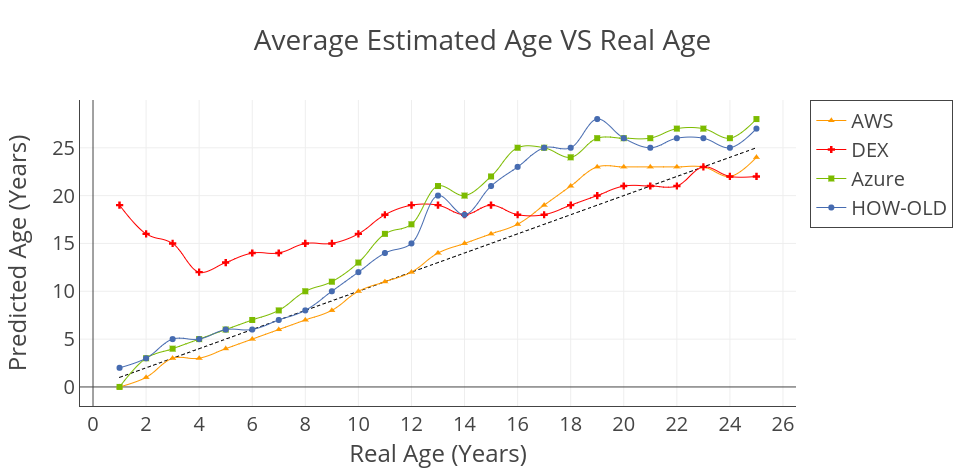}
  \caption{Average Estimated Age from each Service Compared with Actual Age.}
\label{fig:underage_service}
\end{figure*}

\begin{figure*}[h!]
  \centering
  \includegraphics[width=0.82\textwidth,trim={0 0 0 100 cm},clip]{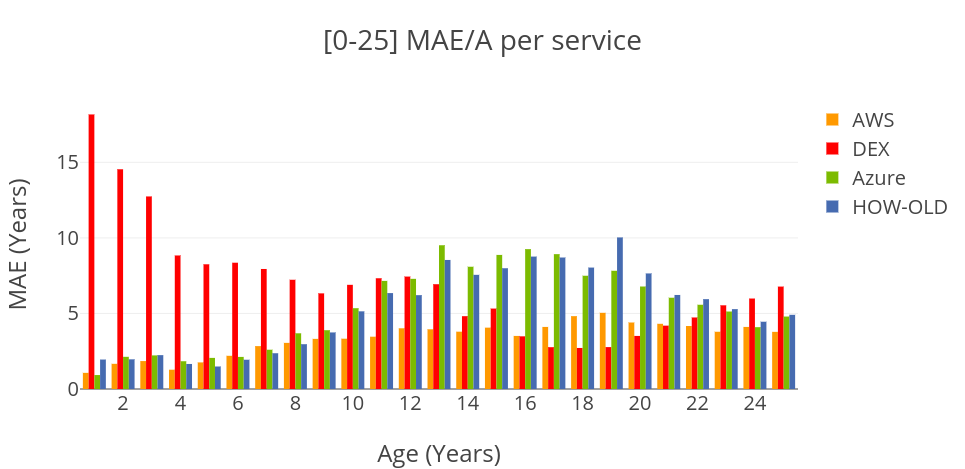}
  \caption{Mean Absolute Error per Age by Service.}
\label{fig:bar_plot}
\end{figure*}

The evaluation for the first experiment focused on samples from 0 to 25. The results of the evaluation are shown in Figure~\ref{fig:underage_service}, with the average predicted age for each service plotted against the subjects' actual ages. The MAE for each service can be seen in Figure~\ref{fig:bar_plot} and the average MAE for underage subjects is presented in Table~\ref{table:mae_underage}.

From these figures, it can be seen that Amazon Rekognition performs best overall. Although it has a slight tendency towards underestimation up to the age of 12, it maintains its accuracy in older age groups better than Azure and How-Old.net, whose predictions gradually deviate away from the real age between the ages of 10 and 22. These three services show similar accuracy for the youngest subjects below the age of 12.

In contrast, DEX's pretrained model fails to accurately classify the younger samples. However, from 17 to 21 years old (in the crucial underage/adulthood boundary zone), it has a better performance than the rest of the models. This pattern is likely due to a lack of sufficient sample images used to train the Deep Expectation model for very young subjects, and is the primary reason why DEX's overall MAE is higher than the others.

In terms of overall MAE for underage subjects, the AWS biometric detector service performed better than the rest of the services with a MAE of 
3.347 as shown in Table~\ref{table:mae_underage}. Although the output of the prediction accomplished by AWS was classified with a high and low range, we found that the closest value to the real age would be the lowest value. AWS's superiority is unrivalled across the majority of age ranges, in fact it is between the best two performers for each age. It is also observed that only DEX and AWS underestimated the subjects' ages at any point, while the remaining services overestimated the values almost throughout the entire age range.

\begin{table}[h!]
\centering
\begin{tabular}{|c | c|} 
 \hline
 \textbf{Service} & \textbf{MAE} \\ [0.5ex] 
 \hline\hline\hline
Amazon Rekognition	&	3.349		\\
How-Old.net			&	5.281		\\
Microsoft Azure		&	5.347		\\
(D)eep (EX)pectation&	6.936		\\	[1ex]
 \hline
\end{tabular}
\caption{Mean Absolute Error for Underage Images per Service.}
\label{table:mae_underage}
\end{table}

\subsection{Development of a Deep Learning Model for Age Estimation (DS13K)}
\label{sec:exp2}

The previously-mentioned DEX model in Section~\ref{existing} was built on a VGG-16 architecture. For the development of our model, transfer learning was used; our DS13K model was fine-tuned on DEX in order to take advantage of the preexisting layer weights. Furthermore, the 12,792 images used for training and testing (80\% and 20\% respectively) came from sources described in Section~\ref{curation}. Each input image was resized to a dimension of 224 x 224 pixels and the output had a size of 5 (Multi-class classifier) and were mapped to a value pertaining to the following age range classes: [0-5], [6-10], [11-15], [16-17] and [18-25]. The ranges were adapted from the ``Criminal networks involved in the trafficking and exploitation of underage victims in the European Union'' 2018 report\footnote{\url{https://www.europol.europa.eu/publications-documents/criminal-networks-involved-in-trafficking-and-exploitation-of-underage-victims-in-eu}}, which indicates that the classification of subjects into one of these age ranges is sufficient, and that precise age estimation is not crucial for investigators. 

\begin{figure}[h!]
  \centering
  \includegraphics[width=0.35\textwidth]{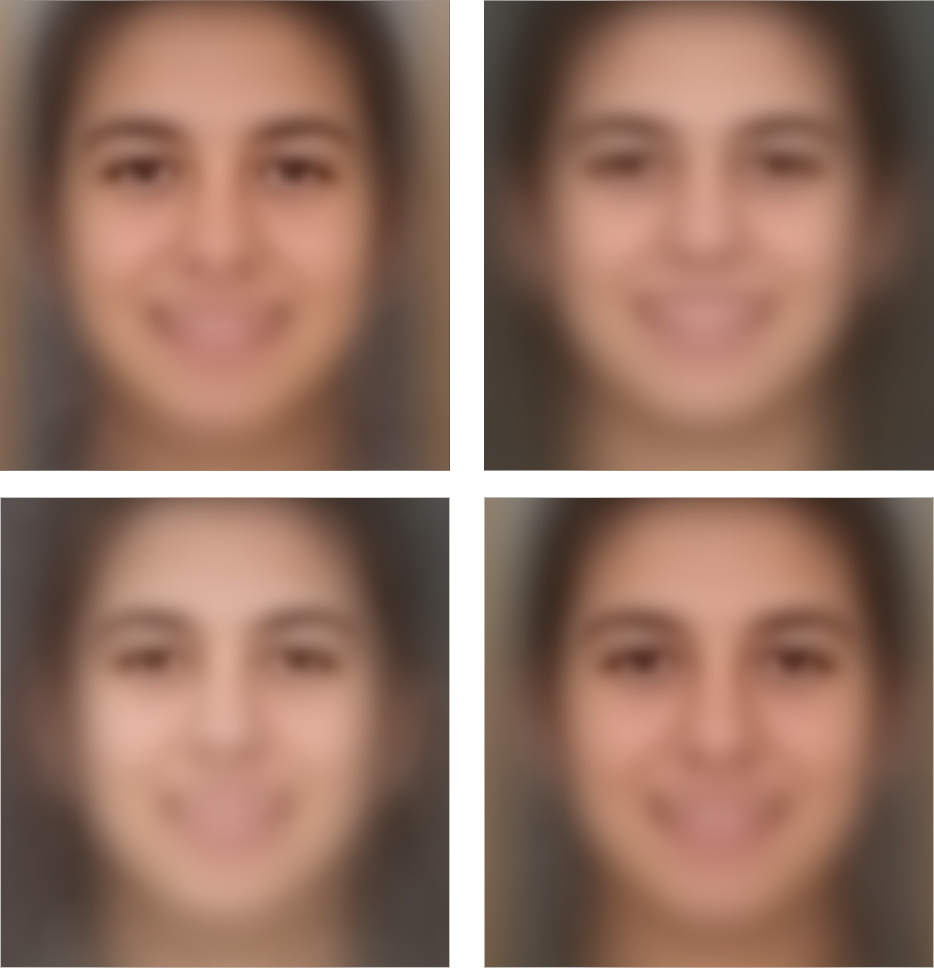}
  \caption{Average Faces of DS13K Subjects between 16 to 17 Years Old.}
\label{fig:average_face}
\end{figure}

To supervise the input of the model, each age class was split into two and the average faces were calculated as depicted in figure \ref{fig:average_face}. The accuracy per age group as well as the average accuracy per service is  in Table~\ref{table:approach1b}, where the best-performing figure for each age range is illustrated in bold. DS13K has the best average performance followed closely by AWS. In the key [16-17] age range, the accuracy of DS13K was substantially higher than the other services, with 68\% of subjects in this range being successfully classified. The second-highest accuracy for this range was AWS with 15\%. As illustrated previously in Figure~\ref{fig:underage_service}, all the other services tend to overestimate age for subjects in this range, which would lead to underage victims being classified as adults. This overestimation of age is also the primary reason why the accuracy in the top age range [18-25] is higher for these services.

\begin{table}[h!]
\centering
\begin{tabular}{|c|c|c|c|c|c|} 
 \hline
\textbf{Range} & \textbf{AWS} & \textbf{Azure} & \textbf{DEX} & \textbf{DS13K} & \textbf{How Old} \\ 
&&&& (our approach) & \\
 \hline\hline\
0-5		&	\textbf{0.88} & 0.69 &	0.00 & 0.77	&	0.78	\\
6-10	&	0.43 & \textbf{0.66} &	0.13 & 0.44	&	0.49	\\
11-15	&	\textbf{0.40} & 0.15 &	0.25 & 0.16	&	0.24	\\
16-17	&	0.15 & 0.00 &	0.17 & \textbf{0.68}	&	0.03	\\
18-25	&	0.87 & \textbf{0.97} &	0.89 & 0.70	&	0.95	\\
\hline
\hline
AVG	&	0.550	&	0.496	&	0.293	&	\textbf{0.553}	&	0.503\\	[1ex]
\hline
\end{tabular}
\caption{Accuracy per Group per Service.}
\label{table:approach1b}
\end{table}


Due to the results encountered by our proposed model, and promising figures for an age range which is of interest to us because of its proximity to the borderline of adulthood [16-17], we decided to include the model in the ensemble approach experiment discussed in the next section.

\subsection{Comparison with Ensemble Learning Techniques.} \label{sec:exp3}

The third experiment was intended to investigate whether Machine Learning (ML) ensemble techniques can be used to improve on the performance exhibited by the existing systems beyond that of each individually. Ensemble techniques are generally defined as those that combine the results of several individual ML algorithms. Given that the existing systems all rely on ML technology, any combination of their results constitutes an ensemble approach. Because the aim of the activity is to compute a predicted age for each subject, regression techniques were considered for this task.

Three standard regression techniques were chosen, namely a logistic regression, gradient boosting and a bagging regressor. These were chosen after observing the results of a number of other regression techniques on this problem. To calculate predicted ages for all of the subjects in the dataset, 10-fold cross validation was used. Here, 90\% of the dataset is used for training, with the regressors tasked with predicting ages for the remaining 10\%. The training data consisted of the predicted ages for each subject image provided by five systems: AWS, How-Old.net, Azure, DEX and DS13K. This process is repeated 10 times so that the predictions are computed for the entire dataset.

To evaluate this experiment, the results of the regression output were compared to each of the five input systems. This comparison was conducted in two ways: firstly the overall MAE was calculated for each technique, and following this the classification accuracy was calculated for the same age ranges used in the previous section. The MAE for each technique across the entire age range [0-25]  is shown in Table~\ref{tab:0-25_MAE}.

\begin{table}[!htb]
\centering
\begin{tabular}{|c|c|}
\hline
\textbf{Method} & \textbf{MAE} \\
\hline\hline\hline

\textbf{GradientBoostingRegressor}  &   \textbf{2.425} \\
\textbf{BaggingRegressor}           &	\textbf{2.623} \\
\textbf{LogisticRegression}         &	\textbf{3.120} \\
AWS                          		&	3.349 \\
DS13K                               &   3.964 \\
How-Old.net                         &   5.281 \\
Azure                        		&	5.347 \\
DEX                          		&	6.936 \\

\hline
\end{tabular}
\caption{Mean Absolute Error Rates for the 0-25 Age Range.}
\label{tab:0-25_MAE}
\end{table}

This table indicates that the three regression algorithms employed achieve a lower MAE than the individual systems. This is an interesting result in that it demonstrates that the off-the-shelf regression models that were used reduce the age estimation error when compared with the individual systems. This strongly motivates further research into regression techniques as a promising method to reducing error rates for the facial age estimation problem. Given that the various systems have different performance characteristics across the age range (as evidenced by the results from Section~\ref{sec:exp1} in particular), these regression models can learn the characteristics of each in order to reduce this effect when combining their outputs.

\begin{figure*}[h!]
  \centering
  \includegraphics[width=0.98\textwidth,trim={0.2cm 0.4cm 0.2cm 53},clip]{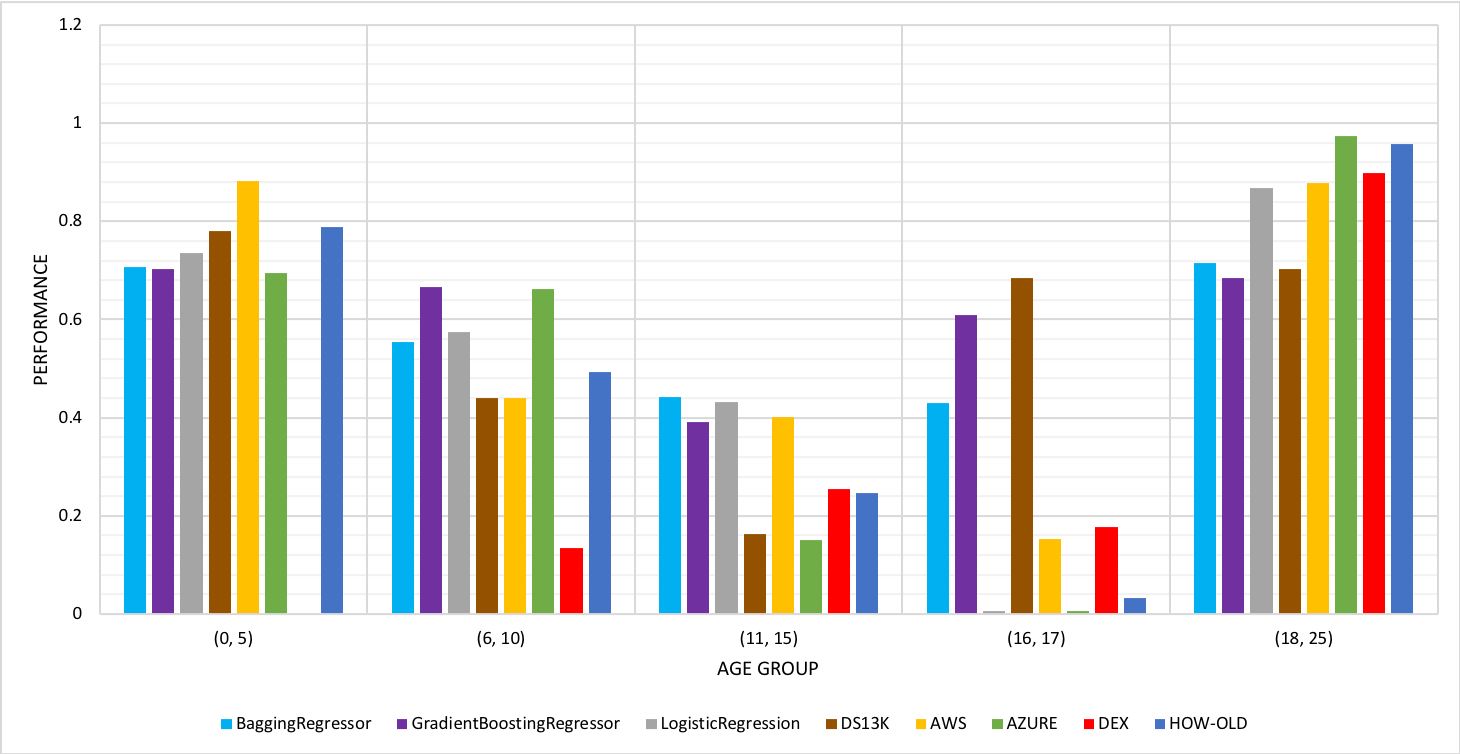}
  \caption{Performance vs Age Group.}
\label{fig:bar_age_range}
\end{figure*}

Given that regression techniques do have a lower error rate than the other approaches within this age range, is it subsequently of interest to find whether their use is also motivated by their performance on the age-range classification task. When the images are divided into age ranges, the accuracy of the regression techniques was also calculated. This did not require a separate experiment to be run; rather an alternative evaluation was conducted. For this evaluation, the important consideration was whether the specific age predicted by the regressor was within the correct age range for each subject. The accuracy of each regressor for each age range is presented in Table~\ref{table:ensemble_approach}, and compared with the underlying input systems in Figure~\ref{fig:bar_age_range}.

\begin{table}[h!]
\centering
\begin{tabular}{|c|c|c|c|} 
 \hline
\textbf{Range} & \textbf{Logistic} & \textbf{Gradient} & \textbf{Bagging} \\
& \textbf{Regression} & \textbf{Boosting} & \textbf{Regressor}\\ [0.5ex] 
 \hline\hline\
0-5	&\textbf{0.734}		&	0.703 & 0.707	\\
6-10 & 0.575	&	\textbf{0.665} &	0.553	\\
11-15 & 0.432	&	0.391 &	\textbf{0.441}	\\
16-17 & 0.006	&	\textbf{0.609} &	0.428	\\
18-25 & \textbf{0.867}	&	0.684 &	0.713	\\	
\hline
\hline
AVG &0.523	& \textbf{0.611}	&0.569\\[1ex]
 \hline
\end{tabular}
\caption{Ensemble Approach Accuracy for Underage Subjects.}
\label{table:ensemble_approach}
\end{table}

From these, it can be seen that the logistic regression, while achieving an overall MAE better than the underlying systems, does not exhibit a promising pattern in terms of the age ranges. Its accuracy in the key 16-17 age range is below almost all other approaches. In contrast, the Gradient Boosting and Bagging approaches both show positive results in this range, with both achieving higher accuracy than the four third-party services that were used.

For underage subjects, the accuracy rates of AWS, How-Old.net and Azure decrease through age ranges as opposed to the adult range [18-25]. It can be observed in Figure~\ref{fig:bar_age_range} that most online services have trouble classifying images in the core [16-17] bracket but that both the Gradient Boosting and Bagging ensemble approaches and the DS13K model have much better accuracy in this range.

Given the results in the previous sections, it is unsurprising that AWS, How-Old.net and Azure have the poorest performance for underage subjects near the borderline. In Section~\ref{sec:exp1}, they are shown to generally overestimate a subject's age in this range, thus frequently misclassifying them as adults. Furthermore, the results in Section~\ref{sec:exp1}, specifically Figure~\ref{fig:bar_plot} indicate that their MAE/Year is greater from the region 13 to 19 years of age in the dataset. Unsurprisingly, the classification accuracy reduces as underage ages get closer to the cut-off point of 18. For 17 year old subjects, DEX's MAE/Year is the lowest, meaning that the performance is better for that particular age than the rest of the services, whereas Azure has the worst performance between them. Their tendency to overestimate ages results in higher accuracy figures for overage subjects. An 18 year old is very rarely (less than 10\% of the time) misclassified as being underage. 

On the other hand, the accuracy of the regression models is much higher than for the underlying systems when averaged over the age ranges. Overall, the Gradient Boosting approach shows the best results. Even for 17 year old subjects, it has a better performance over the rest of ensembles, though failing to beat the DS13K model. 

One notable finding is that the ensemble approaches have lower accuracy for subjects who are equal and over 18. This is partially due to the tendency of the underlying systems to overestimate ages, which will naturally lead to high accuracy for overage subjects in the highest age bracket. However, the accuracy of the regression models for overage subjects is far in excess of the accuracy figures for the underlying systems for underage subjects. This is closely related to their overall lower error rates within this age range.

When evaluating this result, it is also important to keep in mind the use cases for these technologies. Arguably the consequences of misclassifying a younger subject as being overage are much more serious than the opposite scenario. If these systems are to be used in a forensic scenario to automatically identify potential victims of child abuse, it is important that such victims are not missed by these systems. Wrongly classifying a youngster as being older may result in a case not coming to the attention of investigators. In contrast, erroneously allocating an older subject as being younger may ultimately result in wasted investigator effort to examine a situation that is ultimately non-criminal. There is a strong argument to be made that the latter event is much less serious. Even in this scenario, a false positive classification of an adult subject as being underage would trigger a manual evaluation, thus placing investigators in the same position as if the technology was not used.

However, given the multi-year backlog in conducting digital forensic investigations in many jurisdictions~\cite{lillis2016challenges}, clearly an approach that improves accuracy overall is desirable. While the results presented in this section show great promise, it is clear that further work is required to improve the performance of facial age identification even further if it is to be adopted on a wide scale as part of digital forensic investigators' toolkits.

\section{Concluding Remarks}
\label{conclusion}

The four services evaluated in this study where Amazon Rekognition (AWS), Microsoft Azure, Deep Expectation (DEX), and How-Old.net. Initial evaluation results on the age range 0 to 25 years indicated that AWS had the overall lowest error rate, followed by How-Old.net; however, the ages that surround the borderline between minority and adulthood (considered to be 18 for this study) were found to follow a different pattern, where DEX surpassed the performance of AWS and Azure. Furthermore, an additional model named DS13K, based on VGG-16, was trained for this task. This achieved the highest accuracy for the borderline age range (16-17) when compared to the four other systems. Experiments on this dataset indicated that ensemble approaches based on regression substantially outperformed the four systems used for this test, both in terms of mean absolute error and the task of classifying subjects into appropriate age ranges. Gradient Boosting and Bagging Regressor approaches outperformed the best individual system (DEX) for the key borderline range (16-17) by over 40\%. This result offers a strong argument in favour of the proposition that ensemble learning has great potential in improving the precision of facial age determination.

Overall, even off-the-shelf regression techniques have been demonstrated to improve upon the performance of commercial offerings, by combining their outputs effectively. This offers a motivation for further work on bringing AI-based techniques to bear on this and other digital forensic challenges.

\subsection{Future Work}
\label{future}

Our aim is to investigate how to aid digital forensic cases with automated machine learning based techniques. Our objective is to expand this study further through comparative analysis of additional services. We have identified a need for higher-volume datasets for child face recognition to improve our models; once we have collected a dataset with the relevant tags with a considerable size, we would re-train a model specifically for underage images that could help enhance not only age prediction services but also other tools that require identification of child exploitation material.


\bibliographystyle{ACM-Reference-Format}
\bibliography{sample-base}


\begin{thebibliography}{31}


\ifx \showCODEN    \undefined \def \showCODEN     #1{\unskip}     \fi
\ifx \showDOI      \undefined \def \showDOI       #1{#1}\fi
\ifx \showISBNx    \undefined \def \showISBNx     #1{\unskip}     \fi
\ifx \showISBNxiii \undefined \def \showISBNxiii  #1{\unskip}     \fi
\ifx \showISSN     \undefined \def \showISSN      #1{\unskip}     \fi
\ifx \showLCCN     \undefined \def \showLCCN      #1{\unskip}     \fi
\ifx \shownote     \undefined \def \shownote      #1{#1}          \fi
\ifx \showarticletitle \undefined \def \showarticletitle #1{#1}   \fi
\ifx \showURL      \undefined \def \showURL       {\relax}        \fi
\providecommand\bibfield[2]{#2}
\providecommand\bibinfo[2]{#2}
\providecommand\natexlab[1]{#1}
\providecommand\showeprint[2][]{arXiv:#2}

\bibitem[\protect\citeauthoryear{Anda, Lillis, Kanta, Becker, Bou-Harb, Khac,
  and Scanlon}{Anda et~al\mbox{.}}{2019}]%
        {ANDA2019S142}
\bibfield{author}{\bibinfo{person}{Felix Anda}, \bibinfo{person}{David Lillis},
  \bibinfo{person}{Aikaterini Kanta}, \bibinfo{person}{Brett Becker},
  \bibinfo{person}{Elias Bou-Harb}, \bibinfo{person}{Nhien An~Le Khac}, {and}
  \bibinfo{person}{Mark Scanlon}.} \bibinfo{year}{2019}\natexlab{}.
\newblock \showarticletitle{Improving the accuracy of automated facial age
  estimation to aid CSEM investigations}.
\newblock \bibinfo{journal}{\emph{Digital Investigation}}  \bibinfo{volume}{28}
  (\bibinfo{year}{2019}), \bibinfo{pages}{S142}.
\newblock
\showISSN{1742-2876}


\bibitem[\protect\citeauthoryear{Anda, Lillis, Le-Khac, and Scanlon}{Anda
  et~al\mbox{.}}{2018}]%
        {anda2018evaluating}
\bibfield{author}{\bibinfo{person}{Felix Anda}, \bibinfo{person}{David Lillis},
  \bibinfo{person}{Nhien-An Le-Khac}, {and} \bibinfo{person}{Mark Scanlon}.}
  \bibinfo{year}{2018}\natexlab{}.
\newblock \showarticletitle{Evaluating Automated Facial Age Estimation
  Techniques for Digital Forensics}. In \bibinfo{booktitle}{\emph{12th
  International Workshop on Systematic Approaches to Digital Forensics
  Engineering (SADFE), IEEE Security \& Privacy Workshops}}. IEEE.
\newblock


\bibitem[\protect\citeauthoryear{Castrill{\'o}n-Santana, Lorenzo~Navarro, and
  Freire~Obreg{\'o}n}{Castrill{\'o}n-Santana et~al\mbox{.}}{2016}]%
        {castrillon2016boys2men}
\bibfield{author}{\bibinfo{person}{Modesto Castrill{\'o}n-Santana},
  \bibinfo{person}{Jos{\'e}~Javier Lorenzo~Navarro}, {and}
  \bibinfo{person}{Cristina Freire~Obreg{\'o}n}.}
  \bibinfo{year}{2016}\natexlab{}.
\newblock \showarticletitle{Boys2Men, an age estimation dataset with
  applications to detect enfants in pornography content}.
\newblock  (\bibinfo{year}{2016}).
\newblock


\bibitem[\protect\citeauthoryear{Chen, Zhang, Dong, Le, and Rao}{Chen
  et~al\mbox{.}}{2017}]%
        {Chen_2017_CVPR}
\bibfield{author}{\bibinfo{person}{Shixing Chen}, \bibinfo{person}{Caojin
  Zhang}, \bibinfo{person}{Ming Dong}, \bibinfo{person}{Jialiang Le}, {and}
  \bibinfo{person}{Mike Rao}.} \bibinfo{year}{2017}\natexlab{}.
\newblock \showarticletitle{Using Ranking-CNN for Age Estimation}. In
  \bibinfo{booktitle}{\emph{The IEEE Conference on Computer Vision and Pattern
  Recognition (CVPR)}}.
\newblock


\bibitem[\protect\citeauthoryear{Dai, Jin, Xue, Yang, and Yu}{Dai
  et~al\mbox{.}}{2009}]%
        {Dai:2009:EUF:1553374.1553399}
\bibfield{author}{\bibinfo{person}{Wenyuan Dai}, \bibinfo{person}{Ou Jin},
  \bibinfo{person}{Gui-Rong Xue}, \bibinfo{person}{Qiang Yang}, {and}
  \bibinfo{person}{Yong Yu}.} \bibinfo{year}{2009}\natexlab{}.
\newblock \showarticletitle{EigenTransfer: A Unified Framework for Transfer
  Learning}. In \bibinfo{booktitle}{\emph{Proceedings of the 26th Annual
  International Conference on Machine Learning}} \emph{(\bibinfo{series}{ICML
  '09})}. \bibinfo{publisher}{ACM}, \bibinfo{address}{New York, NY, USA},
  \bibinfo{pages}{193--200}.
\newblock
\showISBNx{978-1-60558-516-1}
\urldef\tempurl%
\url{https://doi.org/10.1145/1553374.1553399}
\showDOI{\tempurl}


\bibitem[\protect\citeauthoryear{Dantcheva, Velardo, D'Angelo, and
  Dugelay}{Dantcheva et~al\mbox{.}}{2011}]%
        {Dantcheva2011}
\bibfield{author}{\bibinfo{person}{Antitza Dantcheva}, \bibinfo{person}{Carmelo
  Velardo}, \bibinfo{person}{Angela D'Angelo}, {and} \bibinfo{person}{Jean-Luc
  Dugelay}.} \bibinfo{year}{2011}\natexlab{}.
\newblock \showarticletitle{Bag of soft biometrics for person identification}.
\newblock \bibinfo{journal}{\emph{Multimedia Tools and Applications}}
  \bibinfo{volume}{51}, \bibinfo{number}{2} (\bibinfo{date}{01 Jan}
  \bibinfo{year}{2011}), \bibinfo{pages}{739--777}.
\newblock
\showISSN{1573-7721}
\urldef\tempurl%
\url{https://doi.org/10.1007/s11042-010-0635-7}
\showDOI{\tempurl}


\bibitem[\protect\citeauthoryear{Dong, Liu, and Lian}{Dong
  et~al\mbox{.}}{2016}]%
        {dong2016automatic}
\bibfield{author}{\bibinfo{person}{Yuan Dong}, \bibinfo{person}{Yinan Liu},
  {and} \bibinfo{person}{Shiguo Lian}.} \bibinfo{year}{2016}\natexlab{}.
\newblock \showarticletitle{Automatic age estimation based on deep learning
  algorithm}.
\newblock \bibinfo{journal}{\emph{Neurocomputing}}  \bibinfo{volume}{187}
  (\bibinfo{year}{2016}), \bibinfo{pages}{4--10}.
\newblock


\bibitem[\protect\citeauthoryear{Eidinger, Enbar, and Hassner}{Eidinger
  et~al\mbox{.}}{2014}]%
        {eidinger2014age}
\bibfield{author}{\bibinfo{person}{Eran Eidinger}, \bibinfo{person}{Roee
  Enbar}, {and} \bibinfo{person}{Tal Hassner}.}
  \bibinfo{year}{2014}\natexlab{}.
\newblock \showarticletitle{Age and gender estimation of unfiltered faces}.
\newblock \bibinfo{journal}{\emph{IEEE Transactions on Information Forensics
  and Security}} \bibinfo{volume}{9}, \bibinfo{number}{12}
  (\bibinfo{year}{2014}), \bibinfo{pages}{2170--2179}.
\newblock


\bibitem[\protect\citeauthoryear{Farina, Scanlon, Le-Khac, and Kechadi}{Farina
  et~al\mbox{.}}{2015}]%
        {farina2015overviewcloudforensics}
\bibfield{author}{\bibinfo{person}{Jason Farina}, \bibinfo{person}{Mark
  Scanlon}, \bibinfo{person}{Nhien-An Le-Khac}, {and} \bibinfo{person}{M-Tahar
  Kechadi}.} \bibinfo{year}{2015}\natexlab{}.
\newblock \showarticletitle{{Overview of the Forensic Investigation of Cloud
  Services}}. In \bibinfo{booktitle}{\emph{{10th International Conference on
  Availability, Reliability and Security (ARES 2015)}}}.
  \bibinfo{publisher}{IEEE}, \bibinfo{address}{Toulouse, France},
  \bibinfo{pages}{556--565}.
\newblock
\urldef\tempurl%
\url{https://doi.org/10.1109/ARES.2015.81}
\showDOI{\tempurl}


\bibitem[\protect\citeauthoryear{Ferguson and Wilkinson}{Ferguson and
  Wilkinson}{2017}]%
        {ferguson2017juvenile}
\bibfield{author}{\bibinfo{person}{Eilidh Ferguson} {and}
  \bibinfo{person}{Caroline Wilkinson}.} \bibinfo{year}{2017}\natexlab{}.
\newblock \showarticletitle{Juvenile age estimation from facial images}.
\newblock \bibinfo{journal}{\emph{Science \& Justice}} \bibinfo{volume}{57},
  \bibinfo{number}{1} (\bibinfo{year}{2017}), \bibinfo{pages}{58--62}.
\newblock


\bibitem[\protect\citeauthoryear{Founds, Orlans, Genevieve, and Watson}{Founds
  et~al\mbox{.}}{2011}]%
        {founds2011nist}
\bibfield{author}{\bibinfo{person}{Andrew~P Founds}, \bibinfo{person}{Nick
  Orlans}, \bibinfo{person}{Whiddon Genevieve}, {and} \bibinfo{person}{Craig~I
  Watson}.} \bibinfo{year}{2011}\natexlab{}.
\newblock \showarticletitle{Nist special databse 32-multiple encounter dataset
  ii (meds-ii)}.
\newblock \bibinfo{journal}{\emph{NIST Interagency/Internal Report
  (NISTIR)-7807}} (\bibinfo{year}{2011}).
\newblock


\bibitem[\protect\citeauthoryear{{Fu}, {Guo}, and {Huang}}{{Fu}
  et~al\mbox{.}}{2010}]%
        {5406526}
\bibfield{author}{\bibinfo{person}{Y. {Fu}}, \bibinfo{person}{G. {Guo}}, {and}
  \bibinfo{person}{T.~S. {Huang}}.} \bibinfo{year}{2010}\natexlab{}.
\newblock \showarticletitle{Age Synthesis and Estimation via Faces: A Survey}.
\newblock \bibinfo{journal}{\emph{IEEE Transactions on Pattern Analysis and
  Machine Intelligence}} \bibinfo{volume}{32}, \bibinfo{number}{11}
  (\bibinfo{date}{Nov} \bibinfo{year}{2010}), \bibinfo{pages}{1955--1976}.
\newblock
\showISSN{0162-8828}
\urldef\tempurl%
\url{https://doi.org/10.1109/TPAMI.2010.36}
\showDOI{\tempurl}


\bibitem[\protect\citeauthoryear{Google}{Google}{2018}]%
        {Google2018}
\bibfield{author}{\bibinfo{person}{Google}.} \bibinfo{year}{2018}\natexlab{}.
\newblock \bibinfo{title}{Using AI to help organizations detect and report
  child sexual abuse material online}.
\newblock
\newblock
\urldef\tempurl%
\url{https://www.blog.google/around-the-globe/google-europe/using-ai-help-organizations-detect-and-report-child-sexual-abuse-material-online/}
\showURL{%
\tempurl}


\bibitem[\protect\citeauthoryear{Grd and Ba{\v{c}}a}{Grd and
  Ba{\v{c}}a}{2016}]%
        {grd2016creating}
\bibfield{author}{\bibinfo{person}{Petra Grd} {and} \bibinfo{person}{Miroslav
  Ba{\v{c}}a}.} \bibinfo{year}{2016}\natexlab{}.
\newblock \showarticletitle{Creating a face database for age estimation and
  classification}. In \bibinfo{booktitle}{\emph{Information and Communication
  Technology, Electronics and Microelectronics (MIPRO), 2016 39th International
  Convention on}}. IEEE, \bibinfo{pages}{1371--1374}.
\newblock


\bibitem[\protect\citeauthoryear{Han, Otto, and Jain}{Han
  et~al\mbox{.}}{2013}]%
        {han2013age}
\bibfield{author}{\bibinfo{person}{Hu Han}, \bibinfo{person}{Charles Otto},
  {and} \bibinfo{person}{Anil~K Jain}.} \bibinfo{year}{2013}\natexlab{}.
\newblock \showarticletitle{Age estimation from face images: Human vs. machine
  performance}. In \bibinfo{booktitle}{\emph{2013 International Conference on
  Biometrics (ICB)}}. IEEE, \bibinfo{pages}{1--8}.
\newblock


\bibitem[\protect\citeauthoryear{Kloess, Woodhams, Whittle, Grant, and
  Hamilton-Giachritsis}{Kloess et~al\mbox{.}}{2017}]%
        {kloess2017challenges}
\bibfield{author}{\bibinfo{person}{Juliane~A Kloess}, \bibinfo{person}{Jessica
  Woodhams}, \bibinfo{person}{Helen Whittle}, \bibinfo{person}{Tim Grant},
  {and} \bibinfo{person}{Catherine~E Hamilton-Giachritsis}.}
  \bibinfo{year}{2017}\natexlab{}.
\newblock \showarticletitle{The challenges of identifying and classifying child
  sexual abuse material}.
\newblock \bibinfo{journal}{\emph{Sexual Abuse}} (\bibinfo{year}{2017}),
  \bibinfo{pages}{1079063217724768}.
\newblock


\bibitem[\protect\citeauthoryear{Le, Boydell, Mac~Namee, and Scanlon}{Le
  et~al\mbox{.}}{2018}]%
        {le2018deeplearningmalware}
\bibfield{author}{\bibinfo{person}{Quan Le}, \bibinfo{person}{Ois\'in Boydell},
  \bibinfo{person}{Brian Mac~Namee}, {and} \bibinfo{person}{Mark Scanlon}.}
  \bibinfo{year}{2018}\natexlab{}.
\newblock \showarticletitle{Deep Learning at the Shallow End: Malware
  Classification for Non-Domain Experts}.
\newblock   \bibinfo{volume}{26} (\bibinfo{date}{07} \bibinfo{year}{2018}),
  \bibinfo{pages}{S118 -- S126}.
\newblock
\urldef\tempurl%
\url{https://doi.org/10.1016/j.diin.2018.04.024}
\showDOI{\tempurl}


\bibitem[\protect\citeauthoryear{Levi and Hassner}{Levi and Hassner}{2015}]%
        {levi2015age}
\bibfield{author}{\bibinfo{person}{Gil Levi} {and} \bibinfo{person}{Tal
  Hassner}.} \bibinfo{year}{2015}\natexlab{}.
\newblock \showarticletitle{Age and gender classification using convolutional
  neural networks}. In \bibinfo{booktitle}{\emph{Proceedings of the IEEE
  conference on computer vision and pattern recognition workshops}}.
  \bibinfo{pages}{34--42}.
\newblock


\bibitem[\protect\citeauthoryear{Lillis, Becker, O'Sullivan, and
  Scanlon}{Lillis et~al\mbox{.}}{2016}]%
        {lillis2016challenges}
\bibfield{author}{\bibinfo{person}{David Lillis}, \bibinfo{person}{Brett
  Becker}, \bibinfo{person}{Tadhg O'Sullivan}, {and} \bibinfo{person}{Mark
  Scanlon}.} \bibinfo{year}{2016}\natexlab{}.
\newblock \showarticletitle{{Current Challenges and Future Research Areas for
  Digital Forensic Investigation}}. In \bibinfo{booktitle}{\emph{{The 11th
  ADFSL Conference on Digital Forensics, Security and Law (CDFSL 2016)}}}.
  \bibinfo{publisher}{ADFSL}, \bibinfo{address}{Daytona Beach, FL, USA},
  \bibinfo{pages}{9--20}.
\newblock


\bibitem[\protect\citeauthoryear{Luu, Seshadri, Savvides, Bui, and Suen}{Luu
  et~al\mbox{.}}{2011}]%
        {luu2011contourlet}
\bibfield{author}{\bibinfo{person}{Khoa Luu}, \bibinfo{person}{Keshav
  Seshadri}, \bibinfo{person}{Marios Savvides}, \bibinfo{person}{Tien~D Bui},
  {and} \bibinfo{person}{Ching~Y Suen}.} \bibinfo{year}{2011}\natexlab{}.
\newblock \showarticletitle{Contourlet appearance model for facial age
  estimation}. In \bibinfo{booktitle}{\emph{Biometrics (ijcb), 2011
  international joint conference on}}. IEEE, \bibinfo{pages}{1--8}.
\newblock


\bibitem[\protect\citeauthoryear{Mund}{Mund}{2015}]%
        {mund2015microsoft}
\bibfield{author}{\bibinfo{person}{Sumit Mund}.}
  \bibinfo{year}{2015}\natexlab{}.
\newblock \bibinfo{booktitle}{\emph{Microsoft azure machine learning}}.
\newblock \bibinfo{publisher}{Packt Publishing Ltd}.
\newblock


\bibitem[\protect\citeauthoryear{Phillips, Wechsler, Huang, and Rauss}{Phillips
  et~al\mbox{.}}{1998}]%
        {phillips1998feret}
\bibfield{author}{\bibinfo{person}{P~Jonathon Phillips}, \bibinfo{person}{Harry
  Wechsler}, \bibinfo{person}{Jeffery Huang}, {and} \bibinfo{person}{Patrick~J
  Rauss}.} \bibinfo{year}{1998}\natexlab{}.
\newblock \showarticletitle{The FERET database and evaluation procedure for
  face-recognition algorithms}.
\newblock \bibinfo{journal}{\emph{Image and vision computing}}
  \bibinfo{volume}{16}, \bibinfo{number}{5} (\bibinfo{year}{1998}),
  \bibinfo{pages}{295--306}.
\newblock


\bibitem[\protect\citeauthoryear{Ratnayake, Obertov{\'a}, Dose, Gabriel,
  Br{\"o}ker, Brauckmann, Barkus, Rizgeliene, Tutkuviene, Ritz-Timme,
  et~al\mbox{.}}{Ratnayake et~al\mbox{.}}{2014}]%
        {ratnayake2014juvenile}
\bibfield{author}{\bibinfo{person}{M Ratnayake}, \bibinfo{person}{Z
  Obertov{\'a}}, \bibinfo{person}{M Dose}, \bibinfo{person}{P Gabriel},
  \bibinfo{person}{HM Br{\"o}ker}, \bibinfo{person}{M Brauckmann},
  \bibinfo{person}{A Barkus}, \bibinfo{person}{R Rizgeliene},
  \bibinfo{person}{J Tutkuviene}, \bibinfo{person}{Stefanie Ritz-Timme},
  {et~al\mbox{.}}} \bibinfo{year}{2014}\natexlab{}.
\newblock \showarticletitle{The juvenile face as a suitable age indicator in
  child pornography cases: a pilot study on the reliability of automated and
  visual estimation approaches}.
\newblock \bibinfo{journal}{\emph{International journal of legal medicine}}
  \bibinfo{volume}{128}, \bibinfo{number}{5} (\bibinfo{year}{2014}),
  \bibinfo{pages}{803--808}.
\newblock


\bibitem[\protect\citeauthoryear{Rothe, Timofte, and Gool}{Rothe
  et~al\mbox{.}}{2016}]%
        {Rothe-IJCV-2016}
\bibfield{author}{\bibinfo{person}{Rasmus Rothe}, \bibinfo{person}{Radu
  Timofte}, {and} \bibinfo{person}{Luc~Van Gool}.}
  \bibinfo{year}{2016}\natexlab{}.
\newblock \showarticletitle{Deep expectation of real and apparent age from a
  single image without facial landmarks}.
\newblock \bibinfo{journal}{\emph{International Journal of Computer Vision
  (IJCV)}} (\bibinfo{date}{July} \bibinfo{year}{2016}).
\newblock


\bibitem[\protect\citeauthoryear{Seiffert, Khoshgoftaar, Van~Hulse, and
  Napolitano}{Seiffert et~al\mbox{.}}{2008}]%
        {seiffert2008rusboost}
\bibfield{author}{\bibinfo{person}{Chris Seiffert}, \bibinfo{person}{Taghi~M
  Khoshgoftaar}, \bibinfo{person}{Jason Van~Hulse}, {and} \bibinfo{person}{Amri
  Napolitano}.} \bibinfo{year}{2008}\natexlab{}.
\newblock \showarticletitle{RUSBoost: Improving classification performance when
  training data is skewed}. In \bibinfo{booktitle}{\emph{Pattern Recognition,
  2008. ICPR 2008. 19th International Conference on}}. IEEE,
  \bibinfo{pages}{1--4}.
\newblock


\bibitem[\protect\citeauthoryear{Thomee, Shamma, Friedland, Elizalde, Ni,
  Poland, Borth, and Li}{Thomee et~al\mbox{.}}{2016}]%
        {thomee2016yfcc100m}
\bibfield{author}{\bibinfo{person}{Bart Thomee}, \bibinfo{person}{David~A
  Shamma}, \bibinfo{person}{Gerald Friedland}, \bibinfo{person}{Benjamin
  Elizalde}, \bibinfo{person}{Karl Ni}, \bibinfo{person}{Douglas Poland},
  \bibinfo{person}{Damian Borth}, {and} \bibinfo{person}{Li-Jia Li}.}
  \bibinfo{year}{2016}\natexlab{}.
\newblock \showarticletitle{YFCC100M: The new data in multimedia research}.
\newblock \bibinfo{journal}{\emph{Commun. ACM}} \bibinfo{volume}{59},
  \bibinfo{number}{2} (\bibinfo{year}{2016}), \bibinfo{pages}{64--73}.
\newblock


\bibitem[\protect\citeauthoryear{Wallhoff}{Wallhoff}{2006}]%
        {fgnet}
\bibfield{author}{\bibinfo{person}{Frank Wallhoff}.}
  \bibinfo{year}{2006}\natexlab{}.
\newblock \bibinfo{title}{{Facial Expressions and Emotions Database}}.
\newblock
\newblock
\urldef\tempurl%
\url{http://www-prima.inrialpes.fr/FGnet/html/home.html}
\showURL{%
\tempurl}


\bibitem[\protect\citeauthoryear{Wang}{Wang}{2003}]%
        {wang2003artificial}
\bibfield{author}{\bibinfo{person}{Sun-Chong Wang}.}
  \bibinfo{year}{2003}\natexlab{}.
\newblock \showarticletitle{Artificial neural network}.
\newblock In \bibinfo{booktitle}{\emph{Interdisciplinary computing in java
  programming}}. \bibinfo{publisher}{Springer}, \bibinfo{pages}{81--100}.
\newblock


\bibitem[\protect\citeauthoryear{Watch}{Watch}{2010}]%
        {watch2010us}
\bibfield{author}{\bibinfo{person}{Economy Watch}.}
  \bibinfo{year}{2010}\natexlab{}.
\newblock \showarticletitle{US Economy}.
\newblock \bibinfo{journal}{\emph{Economy Watch}} (\bibinfo{year}{2010}).
\newblock


\bibitem[\protect\citeauthoryear{Weber, {Cruz Rodrigues}, and Mateus}{Weber
  et~al\mbox{.}}{2016}]%
        {Weber2016}
\bibfield{author}{\bibinfo{person}{Heidi Weber}, \bibinfo{person}{Ant{\'{o}}nio
  {Cruz Rodrigues}}, {and} \bibinfo{person}{Am{\'{e}}rico Mateus}.}
  \bibinfo{year}{2016}\natexlab{}.
\newblock \showarticletitle{{Emotion and Mood in Design Thinking}}.
\newblock \bibinfo{journal}{\emph{Design Doctoral Conference'16: TRANSversality
  - Proceedings of the DDC 3rd Conference}} \bibinfo{number}{July}
  (\bibinfo{year}{2016}), \bibinfo{pages}{65--72}.
\newblock


\bibitem[\protect\citeauthoryear{Zhang and Qi}{Zhang and Qi}{2017}]%
        {zhifei2017cvpr}
\bibfield{author}{\bibinfo{person}{Song~Yang Zhang, Zhifei} {and}
  \bibinfo{person}{Hairong Qi}.} \bibinfo{year}{2017}\natexlab{}.
\newblock \showarticletitle{Age Progression/Regression by Conditional
  Adversarial Autoencoder}. In \bibinfo{booktitle}{\emph{IEEE Conference on
  Computer Vision and Pattern Recognition (CVPR)}}. IEEE.
\newblock


\end{thebibliography}

\end{document}